\documentclass[letterpaper, 10 pt, conference]{ieeeconf}

\IEEEoverridecommandlockouts                              % This command is only needed if
\usepackage{amsmath} % assumes amsmath package installed
\usepackage{amssymb}  % assumes amsmath package installed
\usepackage{url}
\usepackage{xcolor}
\usepackage{graphicx} % Enhanced support for graphics
\graphicspath{{figures/}}
\usepackage{cite}     % Well-formed numeric citations
\usepackage[caption=false,font=footnotesize]{subfig} % Modern replacement for subfigure

\usepackage{todonotes}

\newcommand{\rev}[1]{\ifhmode\unskip\fi}

\usepackage[breaklinks,colorlinks,citecolor=blue]{hyperref}

\usepackage{booktabs}

\AtBeginDocument{
  \setlength{\abovedisplayskip}{10pt}
  \setlength{\belowdisplayskip}{10pt}
  \setlength{\abovedisplayshortskip}{10pt}
  \setlength{\belowdisplayshortskip}{10pt}
  \setlength{\jot}{8pt} % Extra space between lines in align/gather environments
}

\title{\LARGE \bf
  Ostrich: Taking Large Strides Through Stiff Contact in Differentiable Dynamics
}

\author{Ale\v{s} Ku\v{c}era$^{1}$ and Karel Zimmermann$^{1}$%
  \thanks{$^{1}$The authors are with the Department of Cybernetics, Faculty of
    Electrical Engineering, Czech Technical University in Prague, Prague, Czech Republic.
    {\tt\small kuceral4@fel.cvut.cz}}%
}

\begin{document}

\maketitle
\thispagestyle{empty}
\pagestyle{empty}

%%%%%%%%%%%%%%%%%%%%%%%%%%%%%%%%%%%%%%%%%%%%%%%%%%%%%%%%%%%%%%%%%%%%%%%%%%%%%%%%

\begin{abstract}

  Three properties determine whether a differentiable simulator can drive
  gradient-based optimization through contact: simulation accuracy, gradient
  reliability,~\rev{R1.min} and per-iteration cost. Tape-based engines such as MJX and Newton Semi-Implicit
  require timesteps small enough to keep contacts numerically tractable, and
  their backpropagation memory grows linearly with the number of timesteps $T$.
  Surrogate models bound memory by approximating contact away, but the resulting
  gradients lose the geometry the optimization depends on. We present Ostrich,
  a GPU-accelerated rigid-body simulator that resolves hard contacts and friction
  with non-smooth Newton iteration at large timesteps ($h \sim 10^{-1}\,\text{s}$), and
  differentiates the converged residual via the implicit function theorem,
  reusing the forward Schur complement to compute the adjoint at $\mathcal{O}(1)$
  memory per timestep. On real-robot trajectories over a pallet obstacle,~\rev{R2.2} Ostrich holds MuJoCo's sim-to-real accuracy up to a $50\times$ larger timestep.~\rev{R2.2}\rev{R1.12}\rev{C1} Its
  gradients converge from random initializations where MJX descends slowly and
  Newton Semi-Implicit stalls;~\rev{R1.4} a warm iteration runs $211\times$ faster than MJX's and $4.7\times$ faster than Semi-Implicit's.~\rev{C3} On the same scene Ostrich
  differentiates $8{,}192$ parallel worlds on a single $24\,\text{GB}$ GPU,
  sustaining $29\times$ checkpointed MJX's optimization throughput;~\rev{R2.1} without checkpointing both baselines exhaust memory at far fewer worlds.~\rev{R2.1}\rev{SP} We close with a
  gradient-based trajectory optimization
  demonstration over triangle-mesh terrain across a $10\,\text{s}$ horizon, a setting
  where prior engines either restrict to primitive geometry or face the convergence and memory limits shown above.~\rev{R2.1}\rev{R1.11}

\end{abstract}

%%%%%%%%%%%%%%%%%%%%%%%%%%%%%%%%%%%%%%%%%%%%%%%%%%%%%%%%%%%%%%%%%%%%%%%%%%%%%%%%

\section{INTRODUCTION}
\label{sec:intro}

Differentiable physics simulators promise gradient-based trajectory optimization and
policy learning in robotics~\cite{newbury2024review, degrave2019differentiable, xu2022shac}. On long-horizon tasks with stiff, persistent contact, such
as a wheeled robot driving across high-friction terrain, current engines force a
tradeoff between simulation accuracy, gradient reliability,~\rev{R1.min} and per-iteration cost.

\begin{figure}[t]
  \vspace*{2mm}
  \centering
  \includegraphics[width=\columnwidth]{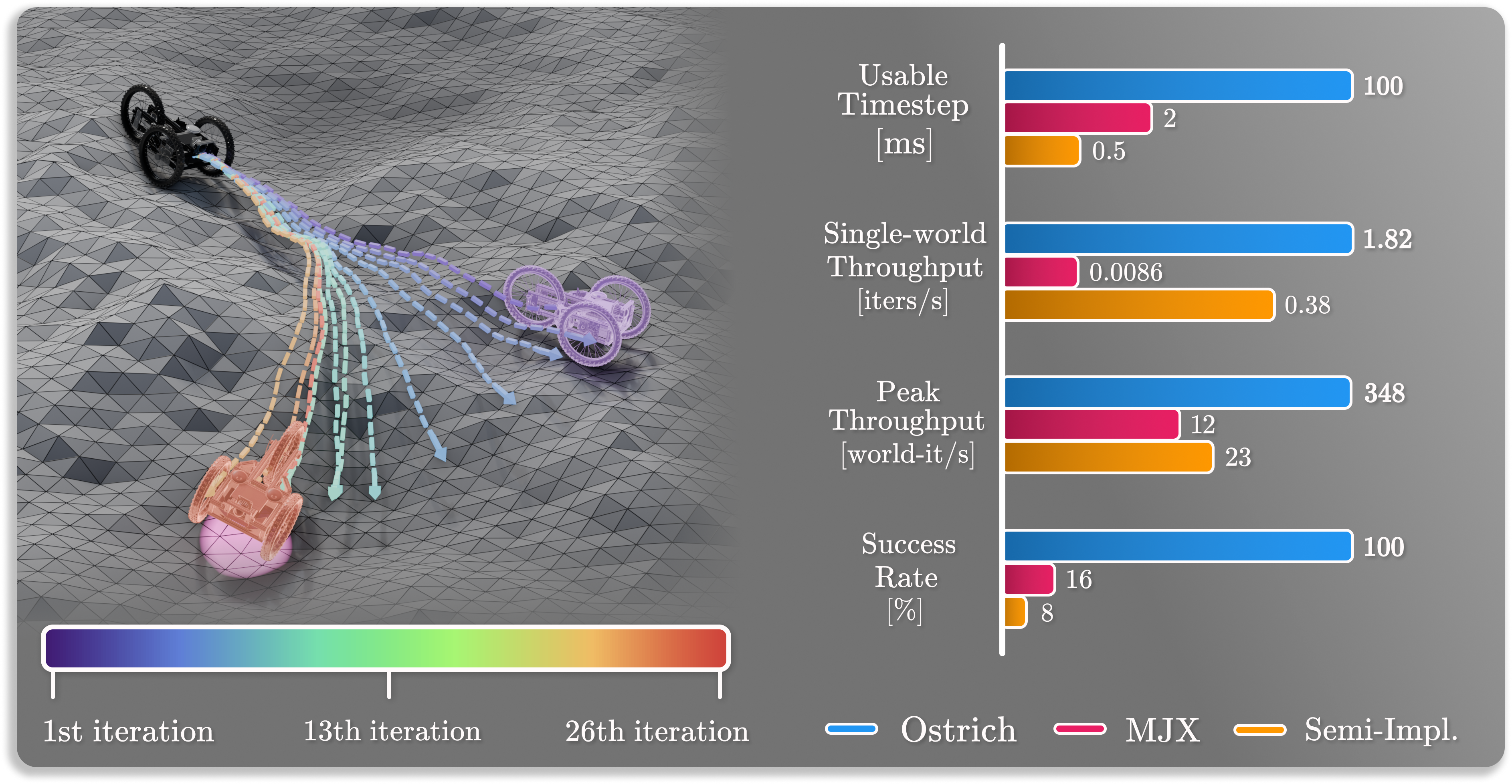}
  \caption{%
    \textbf{Gradient-based trajectory optimization through mesh terrain with
    Ostrich.} \emph{Left:} a wheeled robot's control spline is optimized to
    reach the target (pink area) across a triangle-mesh heightmap. Trajectories
    are colored by optimization iteration, from the initial guess (purple) to
    the converged solution (red) after $26$ iterations; each iteration
    backpropagates through the mesh-wheel contact forces (contact geometry held fixed within each step, \S\ref{sec:terrain}) at
    $\mathcal{O}(1)$ memory per timestep.~\rev{R1.11} \emph{Right:} headline comparison
    against MJX and Newton Semi-Implicit on the box-traversal scene
    (\S\ref{sec:experiments}): usable timestep, single-world optimization throughput (\S\ref{sec:gradient}), peak optimization throughput across parallel worlds, and
    control-synthesis success rate, on a single $24\,\text{GB}$ GPU. The first
    three axes use a log scale.~\rev{R2.1}\rev{C1}\rev{C3}
  }
  \label{fig:hero}
\end{figure}

Tape-based engines need timesteps small enough to keep contacts tractable, and memory linear in step count~\cite{hu2020difftaichi, qiao2021efficient}; surrogate models bound memory but lose the geometry of\rev{SP} non-holonomic traction, localized slip, and mesh-level friction~\cite{corl2024locomotion, yang2025contactsdf}. Implicit-differentiation engines~\rev{R1.5} with
hard contact formulations avoid both pitfalls in principle, yet remain CPU-bound or
restricted to primitive collision geometries~\cite{howelldojo2022, nimble2021,
tinydiffsim}.

We introduce \textbf{Ostrich} (Fig.~\ref{fig:hero}), a GPU-accelerated differentiable
rigid-body simulator that combines hard contacts with $\mathcal{O}(1)$
gradient memory per timestep (total $\mathcal{O}(T)$ stored states over a
horizon, reducible by checkpointing, \S\ref{sec:scalability}).~\rev{R1.6} The forward
integrator is backward-Euler, with hard Non-Linear Complementarity Problem (NCP)
contacts and Coulomb friction resolved inside each step by a non-smooth Newton solver.
This yields stable simulation at macroscopic timesteps ($h \sim 10^{-1}$\,s).
The backward pass applies the Implicit Function Theorem (IFT) to the converged residual,
reusing the forward Schur complement to compute reverse-mode gradients~\rev{R1.5} via a single
adjoint solve, decoupling gradient memory from the solver's internal iterations.

\rev{SP}Beyond the non-smooth Newton formulation of Macklin et
al.~\cite{macklin2019nonsmooth}, we add control constraints solved jointly with contact and well posed at any positive gain\rev{R2.3}, best-iterate backtracking, and robust friction
projection, and we scale both the forward solve and the adjoint to arbitrary triangle meshes and to thousands of parallel worlds.~\rev{R1.gen}
Section~\ref{sec:experiments} evaluates Ostrich against MJX and NVIDIA Newton
Semi-Implicit, the two surveyed engines that run our scene as differentiable baselines, on sim-to-real accuracy,
gradient-based control synthesis, and world-scaling throughput.~\rev{SP}\rev{R1.4}

\section{RELATED WORK}
\label{sec:related_work}

See~\cite{newbury2024review} for a recent survey of differentiable rigid-body
simulators. Three properties motivate Ostrich: simulation accuracy under stiff
contact, gradient reliability,~\rev{R1.min} and per-iteration cost; we organize
prior work by the one each approach sacrifices.~\rev{SP}

\subsection{Simulation Accuracy under Stiff Contact}

\rev{R1.min}GPU-accelerated engines built on automatic differentiation (AD), such as
MJX~\cite{todorov2012mujoco, mjx2024},~\rev{R1.3}
Brax~\cite{freeman2021brax}, NVIDIA Newton~\cite{newton2025}, and
Genesis~\cite{genesis2025} represent contact through soft constraints, penalty
springs, or convex relaxations.~\rev{SP} These representations
integrate stably under AD, but only at timesteps small enough to keep the smoothed
contact numerically tractable. Stiffer, more accurate contact requires shrinking the step further; relaxing
the stiffness widens the sim-to-real gap~\cite{paulus2025hard}.
ADD~\cite{geilinger2020add} mollifies normal and tangential contact forces,
trading accuracy for gradient smoothness, and adaptive-timestep extensions such
as DiffMJX~\cite{paulus2025hard} reduce accepted steps without breaking the
dependence between contact stiffness and step size.~\rev{SP}

\subsection{Gradient Reliability through Contact}

\rev{R1.min}Differentiating through stiff or non-smooth contact often yields gradients that are
biased, noisy, or even wrong-signed~\cite{suh2022policy, suh2022bundled,
antonova2023rethinking, metz2021gradients}.
AD over a backpropagation-through-time (BPTT)
tape recorded across soft
contacts~\cite{hu2020difftaichi} backpropagates through dynamics that have already
been approximated for solver stability; a recent analysis documents the resulting
instability and even sign flips when contacts are stiff~\cite{paulus2025hard}. SHAC~\cite{xu2022shac} and
PODS~\cite{mora2021pods} truncate rollouts to short horizons, dropping the exact
long-horizon gradient. Others replace the dynamics: Single-Rigid-Body Dynamics (SRBD)
surrogates~\cite{corl2024locomotion} differentiate a simplified floating-base
model, unaware of terrain geometry, non-holonomic traction, and localized
slip, and ContactSDF~\cite{yang2025contactsdf} replaces the contact solve with
a LogSumExp-smoothed closed form, yielding closed-form gradients at the cost of
physically smoothed transitions. Randomized smoothing~\cite{suh2022bundled} instead keeps the simulator intact and averages gradients over noise injections, trading non-smoothness for bias.~\rev{SP}

\subsection{Per-Iteration Cost: Memory and Throughput}

Reverse-mode AD records the full per-step forward computation, including every
inner iteration of the contact solver, accumulating a tape that grows with both
trajectory length and per-step solver work~\cite{hu2020difftaichi}.
Implicit-differentiation frameworks bypass the inner-iteration tape by applying
the IFT at the converged solver state,~\rev{R1.5} giving $\mathcal{O}(1)$ memory per
timestep regardless of solver iteration count.~\rev{SP} Building on early LCP-based differentiable physics~\cite{deavila2018lcp},
Nimble~\cite{nimble2021} computes Jacobians~\rev{R1.5} through hard LCP contact
but remains CPU-bound. Qiao et al.~\cite{qiao2021efficient} bound per-step gradient memory with an adjoint\rev{SP} for articulated bodies, and their DiffSim~\cite{qiao2020diffsim} scales mesh\rev{SP} simulation via localized contact regions, but neither targets the\rev{SP} thousands-of-worlds GPU regime.
TinyDiffSim~\cite{tinydiffsim}, extending the NeuralSim
line~\cite{heiden2021neuralsim}, is GPU-parallel~\rev{R1.5} but restricted to
plane/sphere/capsule primitives.~\rev{SP} Dojo~\cite{howelldojo2022} uses
an exact NCP with second-order cone friction, in the classical constraint-based
tradition~\cite{tasora2016chrono}, but is single-threaded CPU and dormant, with
contact geometry beyond planar surfaces available only as experimental,
manually-wired primitives.~\rev{SP}

\rev{SP}Ostrich brings the hard-contact rigor of Dojo's NCP/IFT approach to the GPU by
building on the non-smooth Newton formulation of Macklin et
al.~\cite{macklin2019nonsmooth},~\rev{SP} with an adjoint backward pass giving reverse-mode gradients~\rev{R1.5} for arbitrary mesh collisions
without the memory explosion of AD, the geometric abstractions of dynamics
surrogates, or the sim-to-real failures of mollified contact.~\rev{SP}\rev{R1.min}

\section{THEORY}
\label{sec:theory}

Our simulator builds upon the non-smooth Newton formulation of Macklin et
al.~\cite{macklin2019nonsmooth}, extended with implicit control constraints,
friction-projection robustness (Section~\ref{sec:constraints}), and best-iterate
backtracking (Section~\ref{sec:newton}).~\rev{SP} Implicit differentiation of a converged
solve is standard~\cite{howelldojo2022, nimble2021}; the contribution is that the adjoint reuses the forward solve's matrix-free Schur operator and preconditioner under hard NCP contact (Section~\ref{sec:adjoint}), so gradients run at the forward solve's parallel efficiency.~\rev{R1.gen}

% ===========================================================================
\subsection{Discrete Equations of Motion}

We represent $n_b$ rigid bodies in maximal
coordinates with generalized positions $\mathbf{q} \in \mathbb{R}^{7n_b}$
(position--quaternion per body) and velocities $\mathbf{u} \in \mathbb{R}^{6n_b}$
(linear--angular per body), related by the kinematic map $\dot{\mathbf{q}} =
\mathbf{G}(\mathbf{q})\,\mathbf{u}$. Discretizing with backward Euler at step size $h$:
\begin{align}
  \mathbf{q}^+ &= \mathbf{q}^- + h\,\mathbf{G}(\mathbf{q}^-)\,\mathbf{u}^+ \label{eq:kin} \\
  \tilde{\mathbf{M}}\,(\mathbf{u}^+ - \mathbf{u}^-) &= h\!\left(\mathbf{f}_\text{ext} +
  \mathbf{J}(\mathbf{q}^+)^\top\boldsymbol{\lambda}^+\right) \label{eq:eom}
\end{align}
where $\tilde{\mathbf{M}} = \mathbf{G}^\top\mathbf{M}\mathbf{G}$ is the generalized mass
matrix, $\mathbf{f}_\text{ext}$ collects gravity and gyroscopic terms,
$\mathbf{J}$ is the stacked constraint Jacobian, and $\boldsymbol{\lambda}^+$ are
the Lagrange multipliers. Here $\mathbf{J}$ is evaluated implicitly at
$\mathbf{q}^+$, which is essential for stable handling of stiff contacts and motors;
$\mathbf{G}$ is evaluated explicitly at $\mathbf{q}^-$ to keep the Karush-Kuhn-Tucker
(KKT) system linear in
$\mathbf{u}^+$ and $\boldsymbol{\lambda}^+$. The velocity update remains fully implicit,
preserving backward Euler's unconditional stability.

% ===========================================================================
\subsection{Constraint Formulation}
\label{sec:constraints}

All physical interactions are encoded as residual equations in $\mathbf{q}^+$,
$\mathbf{u}^+$, $\boldsymbol{\lambda}^+$.

\textbf{Bilateral constraints (joints).} Equality constraints $\mathbf{c}_b(\mathbf{q}^+)
= \mathbf{0}$ with diagonal compliance $\mathbf{E} \geq 0$:
\begin{equation}
  \mathbf{r}_b = \mathbf{c}_b(\mathbf{q}^+) + \mathbf{E}\,\boldsymbol{\lambda}_b^+
  = \mathbf{0}.
  \label{eq:bilateral}
\end{equation}

\textbf{Contact constraints.} Non-penetration is enforced via the Fischer-Burmeister (FB)
NCP function $\phi(a,b) = a + b - \sqrt{a^2+b^2}$, which satisfies
$\phi(a,b)=0 \Leftrightarrow a\geq 0,\,b\geq 0,\,ab=0$:
\begin{equation}
  \mathbf{r}_n = \phi\!\left(\mathbf{c}_n(\mathbf{q}^+),\,\boldsymbol{\lambda}_n^+\right)
  = \mathbf{0}.
  \label{eq:contact}
\end{equation}

\textbf{Friction constraints.} Following~\cite{macklin2019nonsmooth}, we derive friction
from the principle of maximal dissipation. For each contact with tangential Jacobian
$\mathbf{J}_f \in \mathbb{R}^{2 \times n_u}$, the friction multiplier
$\boldsymbol{\lambda}_f^+ \in \mathbb{R}^2$ solves
\begin{equation}
  \min_{\boldsymbol{\lambda}_f^+} \;
  {\mathbf{u}^+}^\top \mathbf{J}_f^\top \boldsymbol{\lambda}_f^+
  \quad \text{s.t.} \quad
  \|\boldsymbol{\lambda}_f^+\| \leq \mu\,\lambda_n^+,
  \label{eq:max_dissipation}
\end{equation}
whose KKT conditions yield complementarity between the slip speed
$v_t = \|\mathbf{J}_f\,\mathbf{u}^+\|$ and the cone margin
$\mu\lambda_n^+ - \|\boldsymbol{\lambda}_f^+\|$:
\begin{equation}
  0 \leq v_t \;\perp\;
  \mu\lambda_n^+ - \|\boldsymbol{\lambda}_f^+\| \geq 0.
  \label{eq:friction_ncp}
\end{equation}
We reformulate \eqref{eq:friction_ncp} via the FB function and derive a compliance weight
$W \geq 0$ that projects forces onto the Coulomb
cone, yielding:
\begin{equation}
  \mathbf{r}_f = \mathbf{J}_f\,\mathbf{u}^+ +
  W\,\boldsymbol{\lambda}_f^+ = \mathbf{0}.
  \label{eq:friction}
\end{equation}
The weight smoothly interpolates between sticking ($W \to 0$, zero slip) and sliding ($W >
0$, force on the cone boundary). Treating $W$ as frozen within each Newton iteration
symmetrizes the friction block by lagging its dependence on the normal impulse
$\lambda_n^+$; this makes the per-iteration Newton step inexact but does not soften the
contact model. At the fixed point the frozen value matches the equilibrium $W$, so the
converged multipliers satisfy the exact Coulomb-cone KKT conditions.

The formulation produces timestep-consistent friction: the converged forces in pure sticking and pure sliding are independent of~$h$.~\rev{SP}
Intermediate Newton iterates can carry physically inadmissible friction forces:
outside the cone, or aligned with the slip direction. This is harmless in the
original formulation~\cite{macklin2019nonsmooth}, which uses only the converged
iterate, but our best-iterate backtracking (\S\ref{sec:newton}) may return an
intermediate one, so each must already be admissible. Two safeguards enforce
this: clamping
$\|\boldsymbol{\lambda}_f^+\|$ to $\mu\lambda_n^+$ keeps the impulse within the cone, and
$W \geq 0$ keeps friction dissipative.~\rev{SP}

\textbf{Control constraints.} Proportional-derivative (PD) motor torques applied as explicit external forces, as
MuJoCo does for position actuators~\cite{todorov2012mujoco}, become unstable when
$h^2 k_p$ exceeds the system's inertia.\footnote{MuJoCo's implicit integrators expand
the forces with respect to velocity only, which stabilizes the damping term but leaves
the position gain explicit.}~\rev{R2.3} Instead, we formulate position and velocity
targets as implicit servo constraints:~\rev{R1.8}
\begin{equation}
  \mathbf{r}_c = \mathbf{J}_c\,\mathbf{u}^+ +
  \alpha\,\boldsymbol{\lambda}_c^+ + \mathbf{b}_c = \mathbf{0},
  \label{eq:control}
\end{equation}
with compliance $\alpha = (k_d + h\,k_p)^{-1}$ and $\mathbf{b}_c$ the joint error divided
by $h$ for a position target, and $\alpha = k_p^{-1}$, $\mathbf{b}_c = -\bar{\mathbf{u}}$ for a target joint velocity $\bar{\mathbf{u}}$; both target rows are evaluated at the current Newton iterate. This compliant form is standard: $\alpha$ is the constraint-force-mixing (CFM) parameter of a soft joint motor in ODE~\cite{ode2006} and Bullet, and the compliance of XPBD~\cite{macklin2016xpbd}. It shares its motivation with Stable PD
control~\cite{tan2011stable}, which also evaluates the servo at the next state; we claim
neither as a contribution. Stable PD predicts the next configuration from the current
velocity, applies the result as an explicit torque with contact held at the current
state, and is stable under $k_d \geq h k_p$~\cite[App.~A]{tan2011stable}. Here
$\boldsymbol{\lambda}_c^+$ is instead a constraint variable solved jointly with contacts
and friction inside the Newton system, well posed for any positive timestep and gains, contact-consistent, and differentiated by the same adjoint.~\rev{R2.3} Wheels are driven in the
velocity-target mode, $\boldsymbol{\lambda}_c^+ = k_p(\omega^{\mathrm{cmd}} - \omega^+)$ with $\omega^{\mathrm{cmd}} = \bar{\mathbf{u}}$ the commanded and $\omega^+ = \mathbf{J}_c\mathbf{u}^+$ the solved wheel speed; no torque saturation is modeled.~\rev{R1.8}

% ===========================================================================
\subsection{Newton Solver}
\label{sec:newton}

Stacking \eqref{eq:kin}--\eqref{eq:control} yields a nonlinear system
$\mathbf{r}(\mathbf{s}^+, \mathbf{s}^-, \mathbf{a}, \boldsymbol{\theta}) = \mathbf{0}$
in the state $\mathbf{s} = [\mathbf{q}, \mathbf{u}, \boldsymbol{\lambda}]^\top$,
where $\mathbf{a}$ are control targets and $\boldsymbol{\theta}$ are physical
parameters. The residual splits into one block per component of $\mathbf{s}$:~\rev{SP}
\begin{align}
  \mathbf{r}_\text{kin} &= \mathbf{q}^+ - \mathbf{q}^-
  - h\,\mathbf{G}(\mathbf{q}^-)\,\mathbf{u}^+, \label{eq:rkin} \\
  \mathbf{r}_\text{dyn} &= \tilde{\mathbf{M}}\,(\mathbf{u}^+ - \mathbf{u}^-)
  - h\!\left(\mathbf{f}_\text{ext} +
  \mathbf{J}^\top\boldsymbol{\lambda}^+\right), \label{eq:rdyn} \\
  \mathbf{r}_\text{con} &=
  \begin{bmatrix}
    \mathbf{r}_b^\top & \mathbf{r}_n^\top & \mathbf{r}_f^\top & \mathbf{r}_c^\top
  \end{bmatrix}^\top, \label{eq:rcon}
\end{align}
where $\mathbf{r}_\text{kin}$ enforces the position update \eqref{eq:kin},
$\mathbf{r}_\text{dyn}$ enforces Newton's second law \eqref{eq:eom}, and
$\mathbf{r}_\text{con}$ stacks all constraint residuals: bilateral joints
\eqref{eq:bilateral}, normal contacts \eqref{eq:contact}, friction
\eqref{eq:friction}, and control \eqref{eq:control}.

We solve this system with an inexact non-smooth Newton
method. Because $\mathbf{G}$ is frozen at
$\mathbf{q}^-$, the kinematic residual \eqref{eq:rkin} is linear in
$\mathbf{q}^+$, so the position is not an independent unknown: setting
$\mathbf{r}_\text{kin} = \mathbf{0}$ gives the substitution
$\Delta\mathbf{q} = h\,\mathbf{G}(\mathbf{q}^-)\,\Delta\mathbf{u}$. We apply it
wherever $\mathbf{q}^+$ enters the other residuals, chiefly through the
constraint functions $\mathbf{c}(\mathbf{q}^+)$, which removes the kinematic
block and leaves a system in $(\Delta\mathbf{u}, \Delta\boldsymbol{\lambda})$
alone. Reducing this KKT block system via the Schur complement then eliminates
$\Delta\mathbf{u}$ and yields:
\begin{equation}
  \underbrace{\left(\mathbf{J}\,\tilde{\mathbf{M}}^{-1}\mathbf{J}^\top +
  \mathbf{C}\right)}_{\mathbf{A}}\,\Delta\boldsymbol{\lambda}
  = \frac{1}{h}\left( \mathbf{J}\,\tilde{\mathbf{M}}^{-1}\,\mathbf{r}_\text{dyn} -
  \mathbf{r}_\text{con}\right),
  \label{eq:schur}
\end{equation}
solved with a matrix-free Preconditioned Conjugate Residual (PCR)
method~\cite{macklin2019nonsmooth}. After solving \eqref{eq:schur} for
$\Delta\boldsymbol{\lambda}$, the velocity step is recovered by back-substitution:
\begin{equation}
  \Delta\mathbf{u} = \tilde{\mathbf{M}}^{-1}(
  \mathbf{J}^\top\Delta\boldsymbol{\lambda}\,h - \mathbf{r}_\text{dyn}).
\end{equation}

\textbf{Best-iterate backtracking.} At large $h$ the Newton iterates can overshoot, so
instead of a line search we run a fixed budget and return the iterate with the smallest
residual $\|\mathbf{r}(\mathbf{s}^k)\|^2$ from index $k_\text{min}$ onward; the floor
$k_\text{min}$ blocks early acceptance of under-converged states.~\rev{SP} On mesh terrain the final iterate overshoots an earlier, lower-residual one by up to three orders of magnitude on a third of steps. The returned iterate is kept cone-admissible by the friction safeguards, so the
timestep-consistency of Section~\ref{sec:constraints} holds to its residual.

% ===========================================================================
\subsection{Adjoint Backward Pass}
\label{sec:adjoint}

Each simulation step defines an implicit mapping
$\mathbf{s}^+ = \mathbf{s}^+(\mathbf{s}^-, \mathbf{a}, \boldsymbol{\theta})$
through the residual equation
$\mathbf{r}(\mathbf{s}^+, \mathbf{s}^-, \mathbf{a}, \boldsymbol{\theta}) = \mathbf{0}$
of Section~\ref{sec:newton}.~\rev{SP}
Given a scalar loss $\mathcal{L}(\mathbf{s}^+)$, the backward pass must propagate the
gradient $d\mathcal{L}/d\mathbf{s}^+$ back to
$d\mathcal{L}/d\mathbf{s}^-$, $d\mathcal{L}/d\mathbf{a}$, and
$d\mathcal{L}/d\boldsymbol{\theta}$. This is a reverse-mode (adjoint-state)
computation, obtained algorithmically via the IFT rather than by
differentiating through the solver's iterations: one backward sweep yields
gradients with respect to all inputs and parameters simultaneously.~\rev{R1.5}\rev{R1.7}

\textbf{Implicit differentiation.}
The chain rule requires the state-transition Jacobian
$d\mathbf{s}^+\!/d\mathbf{s}^-$, which has no closed-form expression since $\mathbf{s}^+$
is defined only implicitly. Differentiating
$\mathbf{r}(\mathbf{s}^+(\mathbf{s}^-,\mathbf{a},\boldsymbol{\theta}),\,
\mathbf{s}^-,\mathbf{a},\boldsymbol{\theta}) = \mathbf{0}$ with respect to $\mathbf{s}^-$
and applying the Implicit Function Theorem yields
\begin{equation}
  \frac{d\mathbf{s}^+}{d\mathbf{s}^-}
  = -\!\left[\frac{\partial\mathbf{r}}{\partial\mathbf{s}^+}\right]^{-1}
  \frac{\partial\mathbf{r}}{\partial\mathbf{s}^-}.
  \label{eq:ift}
\end{equation}
Substituting into the chain rule gives
\begin{equation}
  \frac{d\mathcal{L}}{d\mathbf{s}^-}
  = -\frac{d\mathcal{L}}{d\mathbf{s}^+}
  \left[\frac{\partial\mathbf{r}}{\partial\mathbf{s}^+}\right]^{-1}
  \frac{\partial\mathbf{r}}{\partial\mathbf{s}^-}.
  \label{eq:loss_grad}
\end{equation}

\textbf{Adjoint formulation.}
Rather than forming the dense inverse
$[\partial\mathbf{r}/\partial\mathbf{s}^+]^{-1}$, we introduce adjoint variables
$\mathbf{w}$ satisfying:
\begin{equation}
  \left[\frac{\partial\mathbf{r}}{\partial\mathbf{s}^+}\right]^\top\!\!\mathbf{w} =
  -\nabla_{\mathbf{s}^+}\mathcal{L}.
  \label{eq:adjoint}
\end{equation}
Once $\mathbf{w}$ is solved, all required gradients follow from vector-Jacobian products:
\begin{equation}
  \frac{d\mathcal{L}}{d\boldsymbol{\xi}}
  = \mathbf{w}^\top\frac{\partial\mathbf{r}}{\partial\boldsymbol{\xi}},
  \qquad \boldsymbol{\xi} \in \{\mathbf{s}^-,\, \mathbf{a},\, \boldsymbol{\theta}\}.
  \label{eq:vjp}
\end{equation}

\textbf{Reuse of the forward solver.} Expanding \eqref{eq:adjoint} using the block
structure $\mathbf{w} = [\mathbf{w}_q,\, \mathbf{w}_u,\, \mathbf{w}_\lambda]^\top$ gives
three coupled equations. The first trivially yields
$\mathbf{w}_q = -\nabla_{\mathbf{q}^+}\!\mathcal{L}$, and the third is homogeneous
because the loss does not depend on the Lagrange multipliers directly; substituting into
the remaining equation and eliminating $\mathbf{w}_u$ via the Schur complement yields:
\begin{equation}
  \mathbf{A}\,\mathbf{w}_\lambda =
  -\mathbf{J}\,\tilde{\mathbf{M}}^{-1}\!\left(\nabla_{\mathbf{u}^+}\!\mathcal{L} +
  h\,\mathbf{G}^\top\nabla_{\mathbf{q}^+}\!\mathcal{L}\right),
  \label{eq:adjoint_schur}
\end{equation}
with the \textit{identical} Schur complement matrix $\mathbf{A} = \mathbf{J}\,\tilde{\mathbf{M}}^{-1}\mathbf{J}^\top + \mathbf{C}$ from \eqref{eq:schur}, evaluated at the converged iterate ($W$ at its converged value, its $\lambda_n^+$-dependence lagged as in the forward Jacobian).~\rev{R1.11} After solving for $\mathbf{w}_\lambda$, the velocity
adjoint is recovered via back-substitution:
\begin{equation}
  \mathbf{w}_u = -\tilde{\mathbf{M}}^{-1}\!\left(
    \nabla_{\mathbf{u}^+}\!\mathcal{L}
    + h\,\mathbf{G}^\top\nabla_{\mathbf{q}^+}\!\mathcal{L}
  + \mathbf{J}^\top\mathbf{w}_\lambda\right).
  \label{eq:adjoint_wu}
\end{equation}
The shared $h\,\mathbf{G}^\top\nabla_{\mathbf{q}^+}\!\mathcal{L}$ term is the kinematic substitution \eqref{eq:rkin} transposed.~\rev{SP}
The backward pass therefore requires no additional matrix assembly: only a single PCR
solve with the same operator and preconditioner computed during the forward pass,
achieving $\mathcal{O}(1)$ memory per timestep independent of the forward iteration count.~\rev{R1.11} The full derivative with respect to the previous pose, including the rigid-transform dependence of contact points and normals, is obtained by recording one residual evaluation at the converged state on an AD tape and applying one vector-Jacobian sweep, so
this position pull-back follows the residual by construction rather than being hand-derived.~\rev{R1.2}
Non-smoothness of the Fischer-Burmeister residual at the origin is handled by
$\varepsilon$-smoothed norms ($\varepsilon = 10^{-8}$) and guarded denominators;
the exact non-differentiable point has measure zero and converged iterates do not
sit on it.~\rev{R1.9}

\textbf{Validation.}
We verify the adjoint against centered finite differences on CPU: maximum relative error $0.02\%$ on an impulsive
contact-boundary test, ${\le}0.4\%$ across all wheel DOFs of the three-wheeled robot of \S\ref{sec:experiments} and a four-wheeled model under sticking contact (per-scenario maxima $0.04$--$0.33\%$, central differences with step $10^{-2}$), and
${\le}1.6\%$ on cart-pole losses over horizons up to $50$ steps; the largest relative deviations occur on the DOFs with the smallest gradients; the checks run in continuous integration.~\rev{R1.11}\rev{R1.9}

\section{EXPERIMENTS}
\label{sec:experiments}

\begin{figure*}[t]
  \centering
  \begin{minipage}[c]{0.21\textwidth}
    \includegraphics[width=\linewidth]{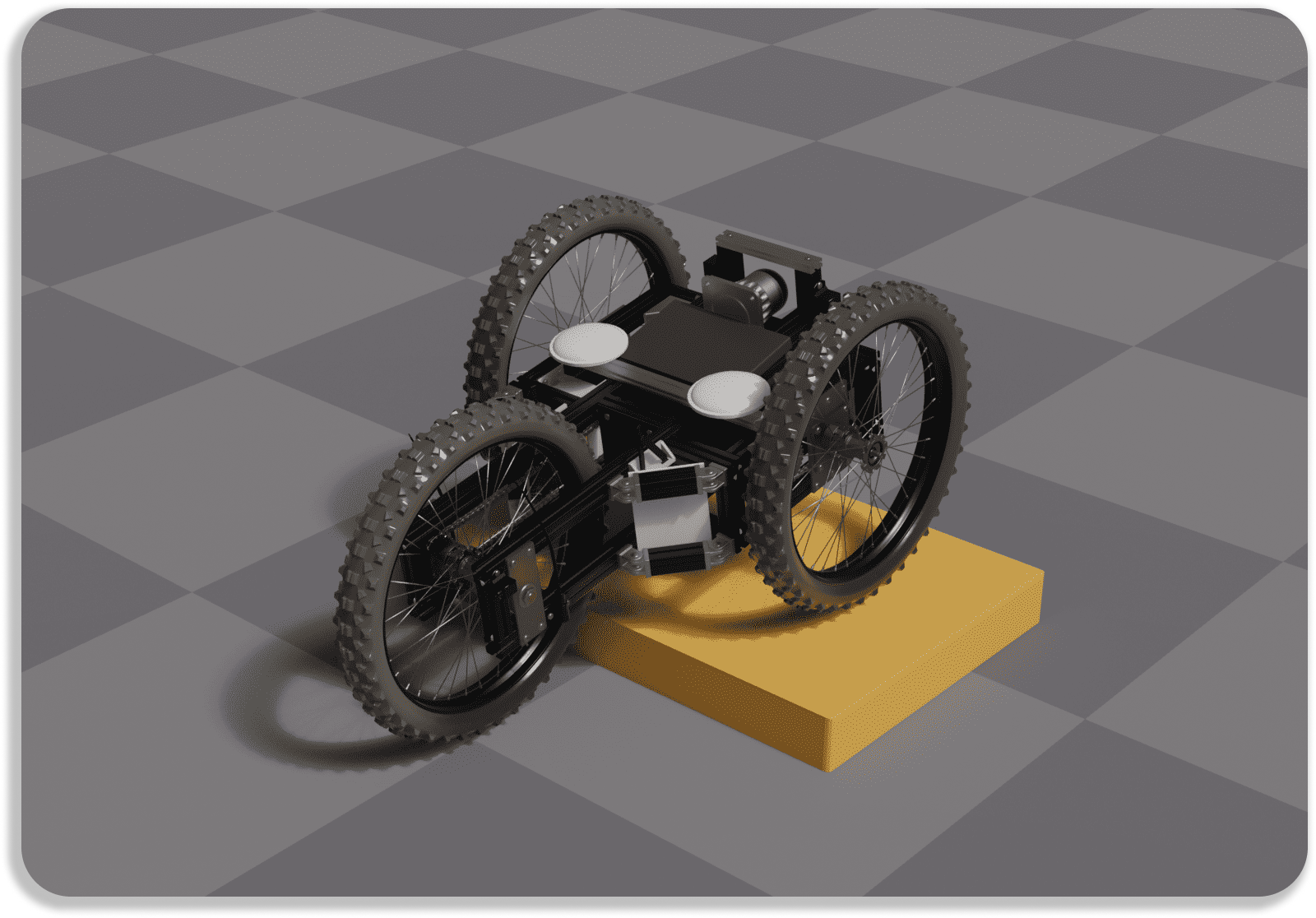}
  \end{minipage}%
  \hspace{0.02\textwidth}%
  \begin{minipage}[c]{0.74\textwidth}
    \includegraphics[width=\linewidth]{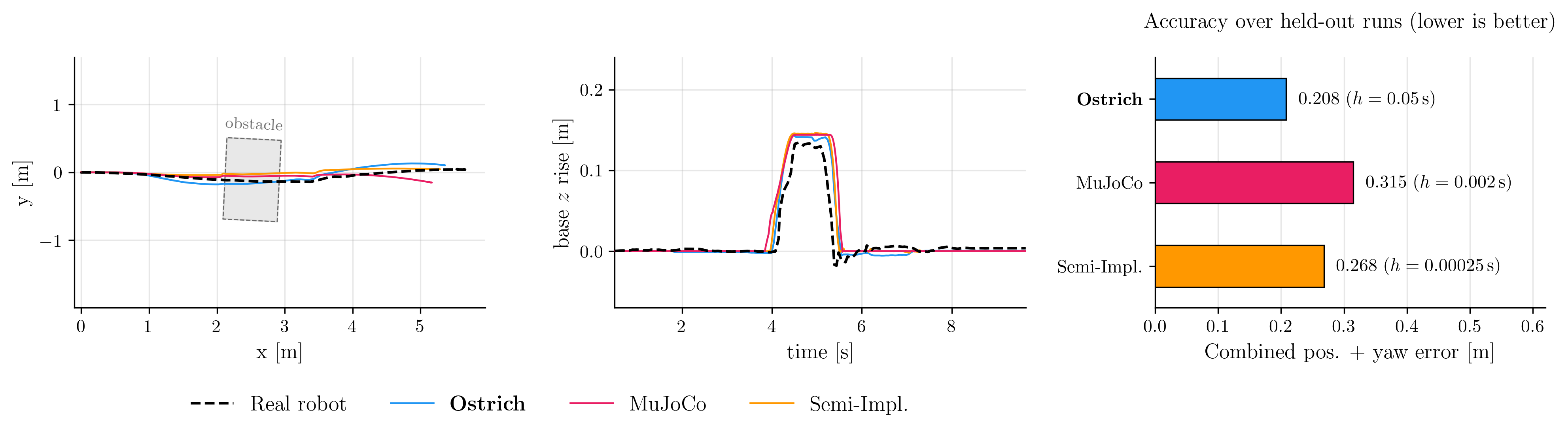}
  \end{minipage}
  \caption{%
    \textbf{Sim-to-real on real pallet traversals (14 runs).} Far left: scene render of the robot mid-climb on the obstacle. Left:
    top-down trajectory for a representative held-out run (selected as the run closest to Ostrich's and MuJoCo's median error), with the per-run fitted
    obstacle footprint in gray and the recorded base path dashed in black.
    Middle: base $z$-elevation; all three engines reproduce the measured
    climb on this held-out run. Right: combined position-and-yaw error averaged over the ten
    held-out runs at each engine's identified parameters.~\rev{R2.2}\rev{SP}
  }
  \label{fig:sim_to_real}
\end{figure*}

We test the three properties in turn: simulation accuracy, gradient
reliability, and cost at the timesteps and parallel scale that applications demand.~\rev{SP}

We evaluate on a custom three-wheeled articulated robot: a $106\,\text{kg}$ vehicle
with $5.5\,\text{kg}$ cylindrical wheels (radius $35\,\text{cm}$,
wheelbase $75\,\text{cm}$) on revolute joints and per-wheel velocity servos. All real-world data come from $14$ traversals of a $14.4\,\text{cm}$ pallet (\S\ref{sec:sim_to_real}).~\rev{R2.2}

Of the six differentiable simulators we surveyed, none produces clean gradients through stiff box contact, and the failure modes track the contact-solver family.
\emph{Penalty} solvers (Newton's Semi-Implicit~\cite{newton2025}, Brax's
\emph{spring} pipeline~\cite{freeman2021brax}) represent contact with stiff compliant springs and so require very small timesteps (\S\ref{sec:sim_to_real}); because backpropagation runs through every stiff step, their gradients are reliable only over short horizons and reach non-finite values over longer ones~\cite{pan2026learning, metz2021gradients}. \emph{Position-based} solvers fare no better:
Newton's \emph{XPBD} only approximates gradients (it disables its position-derived velocity update under AD~\cite{newton2025}),~\rev{SP} and Brax's \emph{positional} pipeline did
not simulate the robot stably on this scene; Brax's \emph{generalized} (QP)
pipeline returns non-finite gradients.~\rev{R1.3}

We therefore benchmark Ostrich against
Newton Semi-Implicit, a maintained, GPU-native representative of the penalty
family, and MJX (\emph{implicitfast} + JAX AD)~\cite{todorov2012mujoco, mjx2024},~\rev{R1.3} whose
convex soft contact is the other widely used approach; \S\ref{sec:gradient} shows both fall short of Ostrich on the box. The remaining engines are excluded outright: Genesis~\cite{genesis2025} produced near-zero rigid-body gradients on this scene, Dojo~\cite{howelldojo2022} is CPU-only and dormant, and TinyDiffSim~\cite{tinydiffsim} supports only primitive geometry.

Ostrich is implemented in NVIDIA Warp~\cite{macklin2022warp} GPU kernels;
Newton~\cite{newton2025} handles model building and collision detection, and
Ostrich adds the non-smooth Newton solver and its adjoint.~\rev{R1.3}
All experiments run on a single NVIDIA RTX~3090 (24\,GB) with an AMD EPYC CPU.

% -----------------------------------------------------------------------
\subsection{Sim-to-Real Accuracy and Timestep Range on a Pallet Obstacle}
\label{sec:sim_to_real}

\rev{R2.2}This section answers two questions on the same real-world dataset:
does each engine's forward pass match held-out recorded trajectories at
identified parameters, and what is the largest $h$ at which it remains both
stable and accurate?~\rev{R2.2}

\paragraph{Setup.}
The dataset comprises $14$ traversals of a wooden pallet
($1.2{\times}0.8\,\text{m}$, $14.4\,\text{cm}$ tall) spanning speeds of $0.12$--$2.05\,\text{m/s}$ and heading changes up to $130^\circ$. Chassis
pose is estimated by onboard lidar-inertial odometry; wheel-velocity
setpoints, logged at $100\,\text{Hz}$, drive each simulator open-loop. The
pallet pose is fitted per run from the lidar clouds and validated against
the climb onset ($\pm3\,\text{cm}$ on all runs).~\rev{R2.2} Error is the yaw-aware
combined metric
$\sqrt{\langle|\Delta p|^2\rangle + (L \cdot \mathrm{RMSE}(\Delta\text{yaw}))^2}$
with lever arm $L{=}0.5\,\text{m}$.~\rev{R1.min} Forward comparisons use reference CPU MuJoCo; its differentiable GPU build, MJX, is used for the gradient and scaling experiments (\S\ref{sec:gradient} onward).~\rev{SP}

\paragraph{Identification protocol.}
Engine parameters are identified on four training runs spanning slow, fast,
and turn-heavy driving, and evaluated on the ten held-out runs; the
wheel-command scale (motor tracking) is calibrated only on the flat
pre-obstacle cruise segment, disjoint from the scored climb. Per engine we
sweep the dominant contact and actuation parameters (Ostrich: servo stiffness $k_p$, lateral wheel friction; MuJoCo: rear-wheel and
torsional friction; Semi-Implicit: penalty stiffness/damping).~\rev{R2.2}\rev{R1.8}

\begin{figure}[t]
  \centering
  \includegraphics[width=\columnwidth]{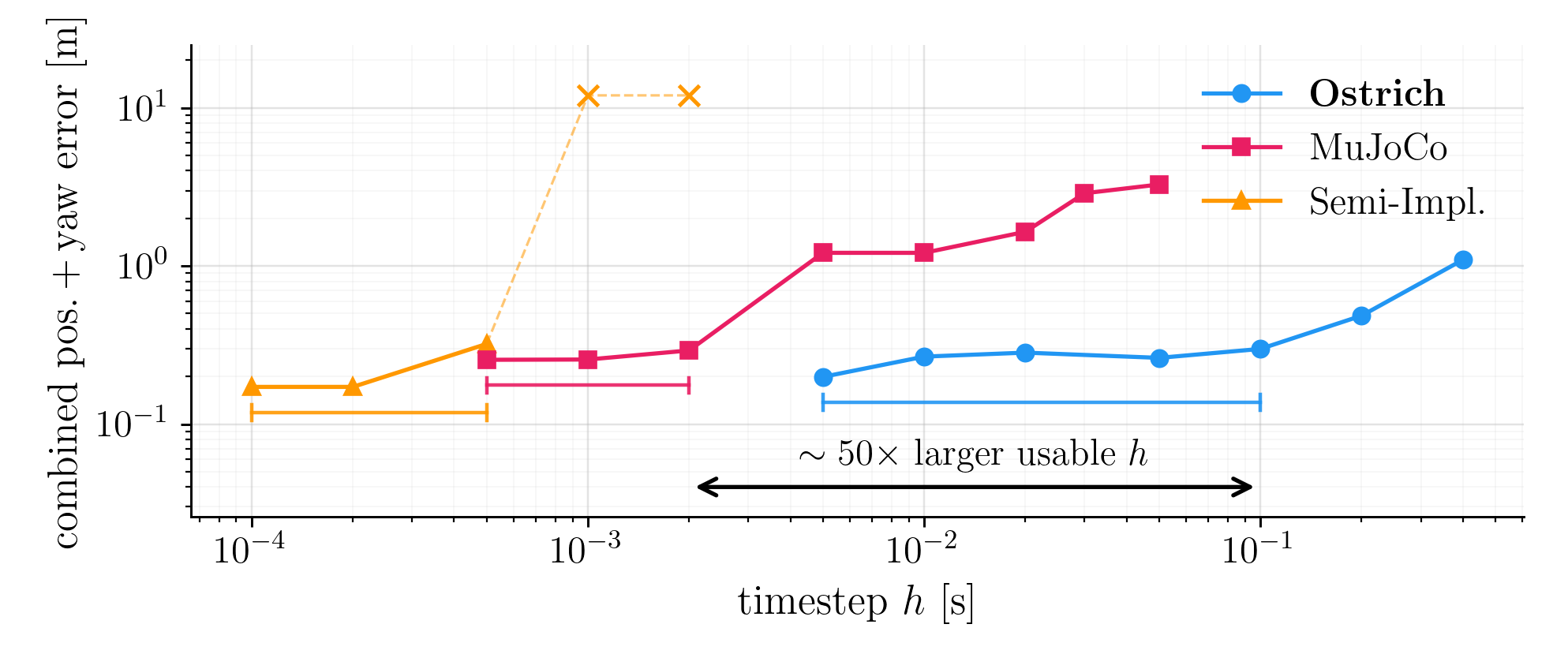}
  \caption{%
    \textbf{Accuracy vs.\ timestep on four held-out runs of the \S\ref{sec:sim_to_real} dataset.}~\rev{R2.2} Each engine runs at its identified parameters over an $h$ grid. Brackets span each engine's usable plateau, where error stays within $2{\times}$ of its sweep minimum (the model-error floor); past the
    edge, integration error takes over. $\times$ marks diverged runs, plotted at the ceiling.~\rev{R1.12} The arrow spans Ostrich's ${\sim}50{\times}$
    larger usable $h$ over MuJoCo (${\sim}200{\times}$ over Semi-Implicit).~\rev{R1.12}
  }
  \label{fig:dt_stability}
\end{figure}

\paragraph{Held-out accuracy.}
Fig.~\ref{fig:sim_to_real} reports the outcome at $h = 50$, $2$ and $0.25\,\text{ms}$ (Ostrich, MuJoCo, Semi-Implicit; each within its Fig.~\ref{fig:dt_stability} plateau): on the held-out runs, Ostrich reaches $0.208\,\text{m}$ combined error, Semi-Implicit
$0.268\,\text{m}$, and MuJoCo $0.315\,\text{m}$, and all three reproduce
the measured pallet climb.~\rev{R2.2} The error is common to all three and concentrates on the turn-heavy runs, implicating the shared friction model
rather than any contact solver: the robot's turn efficiency (yaw response relative to ideal skid steering) varies per run ($0.11$--$0.30$ on four runs, correlated with speed), while a constant-$\mu$ model realizes
exactly one, so every engine's identification saturates at a similar error floor.~\rev{R1.12}\rev{SP} At this floor, simulation accuracy separates the engines by at most $0.11\,\text{m}$; the usable timestep range does by orders of magnitude.~\rev{R2.2}

\paragraph{Timestep range.}
Every gradient experiment that follows pays per step
($T = \text{horizon}/h$), so the decisive property is the largest $h$ at which
an engine stays on that floor. We sweep $h$ at each engine's identified configuration on four held-out runs spanning the speed range; the usable plateau ends where error departs its sweep minimum by more than $2{\times}$ (Fig.~\ref{fig:dt_stability}). Ostrich holds its floor up
to $h = 0.1\,\text{s}$: the implicit hard-contact solve introduces no
contact timescale of its own, and the edge is set by integration error
alone. MuJoCo leaves its floor at a $\mathbf{50{\times}}$ smaller step ($2\,\text{ms}$), where $h$ approaches the contact-softening time constant of its calibration~\cite{todorov2012mujoco}.~\rev{SP}
Semi-Implicit is capped $\mathbf{200{\times}}$ below Ostrich by the explicit-integration stability limit of its penalty springs and diverges beyond $0.5\,\text{ms}$; the held-out evaluation ran it at $0.25\,\text{ms}$, inside that edge.~\rev{R2.2}\rev{R1.12}

% -----------------------------------------------------------------------
\subsection{How Reliable and Fast Are the Gradients?}
\label{sec:gradient}

The forward-side results in \S\ref{sec:sim_to_real} show each engine can
track reality at its identified~\rev{R2.2} parameters, but say nothing about gradient reliability, which we test by task success and optimization convergence (finite-difference agreement: \S\ref{sec:adjoint}).~\rev{R1.min}
We therefore pose a gradient-based control-synthesis task on a simulated
scene that models the \S\ref{sec:sim_to_real} pallet as a rigid box obstacle
(same robot):~\rev{R2.2} recover open-loop wheel commands that drive the robot through
contact to a target pose.~\rev{R2.2}\rev{R1.min}

\paragraph{Setup.}
\rev{R1.3}Per-engine $h$ follows Fig.~\ref{fig:dt_stability}: Ostrich runs at
$\mathbf{100\,\text{ms}}$, the top of its plateau; MJX at $\mathbf{2\,\text{ms}}$, MuJoCo's plateau edge;
Semi-Implicit at $\mathbf{0.5\,\text{ms}}$, the top of its plateau. This gives $60$, $3{,}000$, and $12{,}000$ simulation steps per rollout over the $6\,\text{s}$
horizon.~\rev{C1} The task: from a randomly perturbed initial pose, drive the chassis past the box to a random target pose ($x{=}3\,\text{m}$) and stop there, with no trajectory to imitate.~\rev{SP}
The variable is a $K{=}10$-knot wheel-velocity spline ($30$ parameters);~\rev{R1.1} the loss combines final
position and heading error, a terminal-velocity penalty,~\rev{SP} control regularizers,
and a per-step straight-line tracking term for dense gradient signal. Each
engine runs $50$ Adam iterations on $25$ random (initial-condition, target)
trials from a shared RNG stream.~\rev{SP} Optimizer hyperparameters are swept per engine; all three run at the best learning rate from that sweep ($0.3$, $0.3$, $0.1$ for Ostrich, MJX, Semi-Implicit).~\rev{C2} Solvers: MJX runs MuJoCo 3.9.0's Newton-type constraint solver with its non-differentiable \texttt{while\_loop} replaced by a fixed $10$-iteration scan; Semi-Implicit, symplectic Euler with penalty contacts; Ostrich, backward Euler with $16$ Newton $\times$ $16$ PCR iterations per step; physical parameters per engine as identified in \S\ref{sec:sim_to_real}.~\rev{R1.3}

\begin{figure}[t]
  \centering
  \includegraphics[width=\columnwidth]{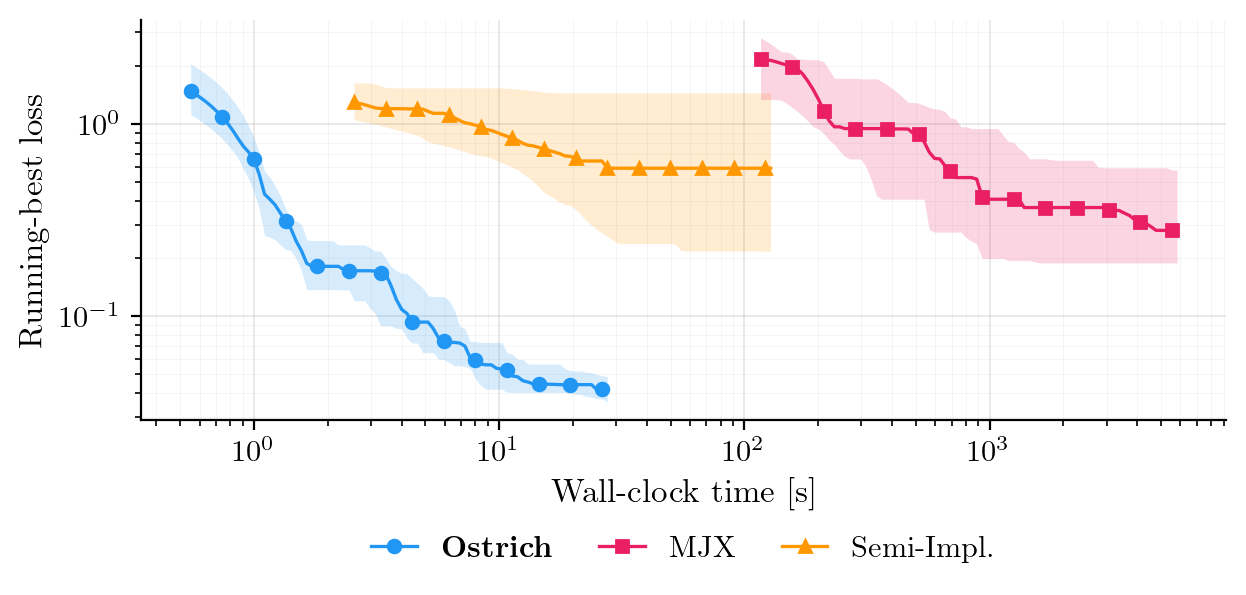}
  \caption{%
    \textbf{Gradient reliability on the box-obstacle control-synthesis task}~\rev{R1.min} (\S\ref{sec:gradient}; $25$ random trials, $50$ Adam iterations per engine).~\rev{SP} Median running-best loss against wall-clock (log $x$) with the interquartile band across trials.~\rev{SP} Ostrich
    converges to a median loss of $0.042$ in ${\sim}28\,\text{s}$ of warm iterations ($29\,\text{s}$ end to end); MJX reaches $0.28$~\rev{C1} only after ${\sim}1.6\,\text{h}$;~\rev{C3} Semi-Implicit stalls at $0.59$.~\rev{R1.11} The $x$-axis excludes each engine's one-time setup (\S\ref{sec:gradient}).~\rev{R1.13}
  }
  \label{fig:gradient_quality}
\end{figure}

\paragraph{Reliability.}
Across the $25$ trials (success: final
position error $<0.2\,\text{m}$ and terminal speed $<0.3\,\text{m/s}$),~\rev{R1.4}
Ostrich succeeds on $\mathbf{100\%}$ of trials at a median final position error of $0.065\,\text{m}$, with a tight cross-trial loss band.~\rev{R1.11}\rev{R1.gen} MJX succeeds on $\mathbf{16\%}$ (median position error $0.32\,\text{m}$) and characteristically
approaches the target without stopping: $8/25$ trials pass the position gate and $9/25$ the speed gate, but only $4/25$ both.~\rev{C1} Semi-Implicit is worse still: its exact reverse-mode gradient is non-finite beyond a ${\sim}2\,\text{s}$ horizon and unreliable below it (non-finite entries are zeroed); salvaging its best finite iterate yields $\mathbf{8\%}$ success ($2/25$)
at $0.58\,\text{m}$ median position error.~\rev{R1.11} Fig.~\ref{fig:gradient_quality} shows the same ordering in optimization loss.~\rev{SP}

\paragraph{Wall-clock cost.}
Ostrich runs one warm (post-setup) optimization iteration in $\mathbf{0.55\,\text{s}}$, $\mathbf{211{\times}}$ faster than MJX ($116\,\text{s}$) and $\mathbf{4.7{\times}}$ faster than Semi-Implicit ($2.6\,\text{s}$), each engine alone on the GPU with setup separated by a two-point fit.~\rev{C3} That setup is not
uniform: $1\,\text{s}$ for Ostrich and $84\,\text{s}$ for MJX, but
$29\,\text{min}$ per trial for Semi-Implicit.~\rev{R1.13} End to end, a $50$-iteration run
costs $\mathbf{29\,\text{s}}$ for Ostrich against $1.6\,\text{h}$ for MJX and
$0.5\,\text{h}$ for Semi-Implicit, and neither baseline reaches a comparable loss.~\rev{C3} The advantage over MJX compounds two factors: $50{\times}$ fewer steps per rollout, and an adjoint that never tapes the inner contact iterations; at scale the gap settles at $29{\times}$ (\S\ref{sec:scalability}).~\rev{R1.6}

\paragraph{Zeroth-order comparison.} We also ran evolution strategies (ES; population $64$ as one batched rollout, $3$ seeds per trial, five trials) on Ostrich's forward pass. On a single problem ES matches first-order in wall-clock: it reaches the success criterion at a median $0.8\times$ the budget of $50$ Adam iterations, with a comparable (marginally lower) loss at every budget up to $10\times$. The difference is sample efficiency. ES spends about $1{,}800$ rollouts per solved problem against $50$ forward-adjoint pairs, a $35\times$ gap that a single world conceals because it leaves the GPU idle (Fig.~\ref{fig:scalability}a); with $64$ worlds, first-order optimizes $64$ independent problems where ES optimizes one. The gap also grows with dimension: at $300$ parameters ES needs twice the rollouts and $1.5\times$ the budget and reaches $4\times$ the matched-budget loss (catching up at $5\times$ the budget), while first-order's cost and outcome do not change.~\rev{R1.1}

% -----------------------------------------------------------------------
\subsection{Does It Scale?}
\label{sec:scalability}

\begin{figure}[t]
  \centering
  \includegraphics[width=\columnwidth]{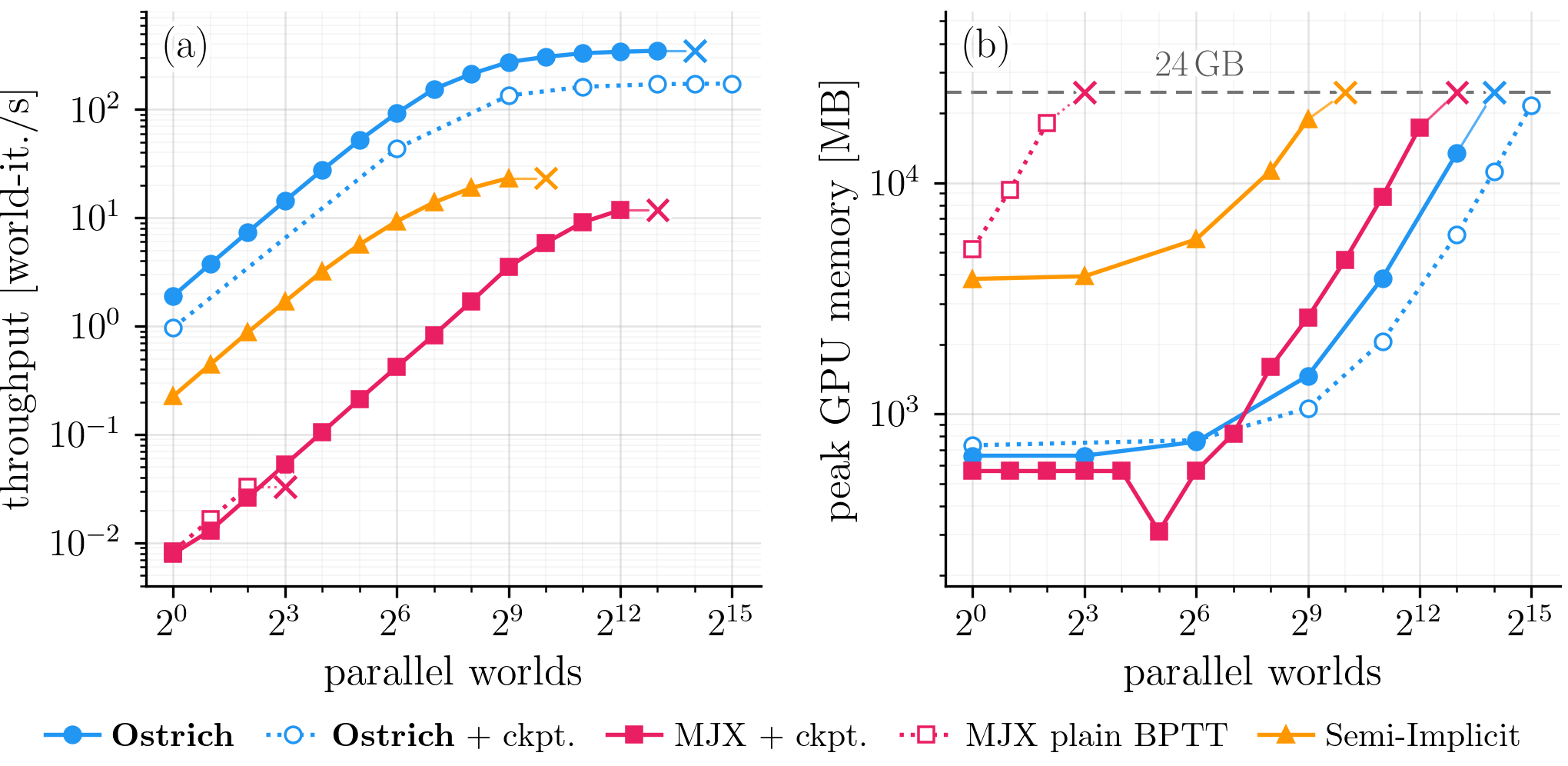}
  \caption{%
    \textbf{Optimization throughput (a) and peak GPU memory (b) vs.\ number
    of worlds} on the box-traversal scene of \S\ref{sec:gradient}
    (RTX 3090, $24\,\text{GB}$; NVML-polled). Solid: each engine's operating
    configuration; dotted: the alternative (checkpointing for Ostrich, plain BPTT for MJX, which needs checkpointing to scale at all). $\times$: first failing batch size. ~\rev{R2.1}~\rev{SP}
  }
  \label{fig:scalability}
\end{figure}

\rev{SP}On the box-traversal scene of \S\ref{sec:gradient}, we measure end-to-end
optimization throughput (forward $+$ backward, in world-iterations per
second) and peak GPU memory (Fig.~\ref{fig:scalability}) across worlds from
$1$ to each engine's capacity, with MJX and Ostrich additionally run under segment checkpointing (bit-exact to the plain gradients), so neither is handicapped by naive gradient storage.\footnote{Memory is the NVML process footprint of the differentiable rollout; forward-only simulation scales far beyond these ceilings. The $24\,$GB card is needed only here; Ostrich's and Semi-Implicit's single-world experiments fit in $4\,$GB, MJX's in $6\,$GB.~\rev{R1.min} A checkpointed Semi-Implicit variant was implemented and validated; its timing is omitted because per-step tape re-recording makes segment recomputation impractical at scale.}~\rev{R2.1}

The baselines' memory pressure has two sources. Semi-Implicit's tiny $h$
forces $12{,}000$ taped steps over the $6\,\text{s}$ horizon; MJX takes
$3{,}000$ but tapes every inner contact iteration, so plain BPTT exhausts the
$24\,\text{GB}$ card at $8$ worlds.~\rev{C1} Ostrich shrinks both factors: $60$ steps at
$h \sim 10^{-1}\,\text{s}$ ($50\times$ fewer than MJX),~\rev{C1} and an adjoint over the
converged residual, so inner iterations are never taped.~\rev{SP}

\paragraph{Throughput across regimes.}
At a single world, the interactive and model-predictive-control (MPC) regime, this measurement gives $0.53\,\text{s}$ per optimization iteration for Ostrich versus $126\,\text{s}$ for checkpointed MJX and $4.4\,\text{s}$ for Semi-Implicit (\S\ref{sec:gradient}'s two-point fit gives $0.55$, $116$ and $2.6\,\text{s}$). At scale, throughput peaks at $\mathbf{348}$ world-iterations/s for Ostrich (at $8{,}192$ worlds) versus $12$ for checkpointed MJX at its $4{,}096$-world memory cap, a $29{\times}$ gap, and $23$ for Semi-Implicit (at $512$; out of memory at $1{,}024$).~\rev{R2.1} Checkpointing changes only where each engine starts paying the recompute cost (one forward re-execution per step): MJX at $8$ worlds, Ostrich at $16{,}384$; Semi-Implicit would at $1{,}024$.~\rev{R2.1} Checkpointed Ostrich sustains $171$--$173$ world-iterations/s up to $\mathbf{32{,}768}$ worlds on $21.7\,\text{GB}$. Computational cost remains
$\mathcal{O}(T)$ for all methods; the gains come from step size and batch
throughput.~\rev{R1.6}\rev{R2.1}

\rev{R2.1}

% -----------------------------------------------------------------------
\subsection{Beyond Primitives: Terrain Traversal}
\label{sec:terrain}

The preceding experiments used a single box obstacle with a few contact pairs. We
now run gradient-based trajectory optimization directly over a triangle mesh
(${\sim}12{,}500$ faces), where mesh-wheel contact normals and friction must be
tracked across hundreds of simultaneous contact candidates per step, the regime
that motivated Ostrich's $\mathcal{O}(1)$-memory backward pass, evaluated on $50$ random instances.~\rev{SP}

\paragraph{Setup.}
Each instance derives from a single seed: a sum-of-sinusoids heightmap ($24 \times 24\,\text{m}$), a random
target spline ($K{=}10$ knots, correlated left/right wheel velocities), and an
initial guess perturbing it with Gaussian noise ($\sigma{=}1.5\,\text{rad/s}$).
The optimizer must recover the target spline's simulated trajectory from that
guess under a position-tracking, yaw-alignment, and control-regularization loss over
$10\,\text{s}$ ($T{=}125$, $h{=}80\,\text{ms}$).~\rev{SP} Gradients flow through the converged Newton solve, including mesh contact and friction forces; on this $10\,\text{s}$ horizon the complete position pull-back of \S\ref{sec:adjoint}, used in full in \S\ref{sec:gradient}, amplifies contact sensitivity (the adjoint norm grows ${\sim}1.13\times$ per step) and gives no usable descent direction, so the adjoint here truncates it, holding contact geometry fixed within each step (cf.~\cite{metz2021gradients}).~\rev{R1.11}

\begin{figure}[t]
  \centering
  \includegraphics[width=\columnwidth]{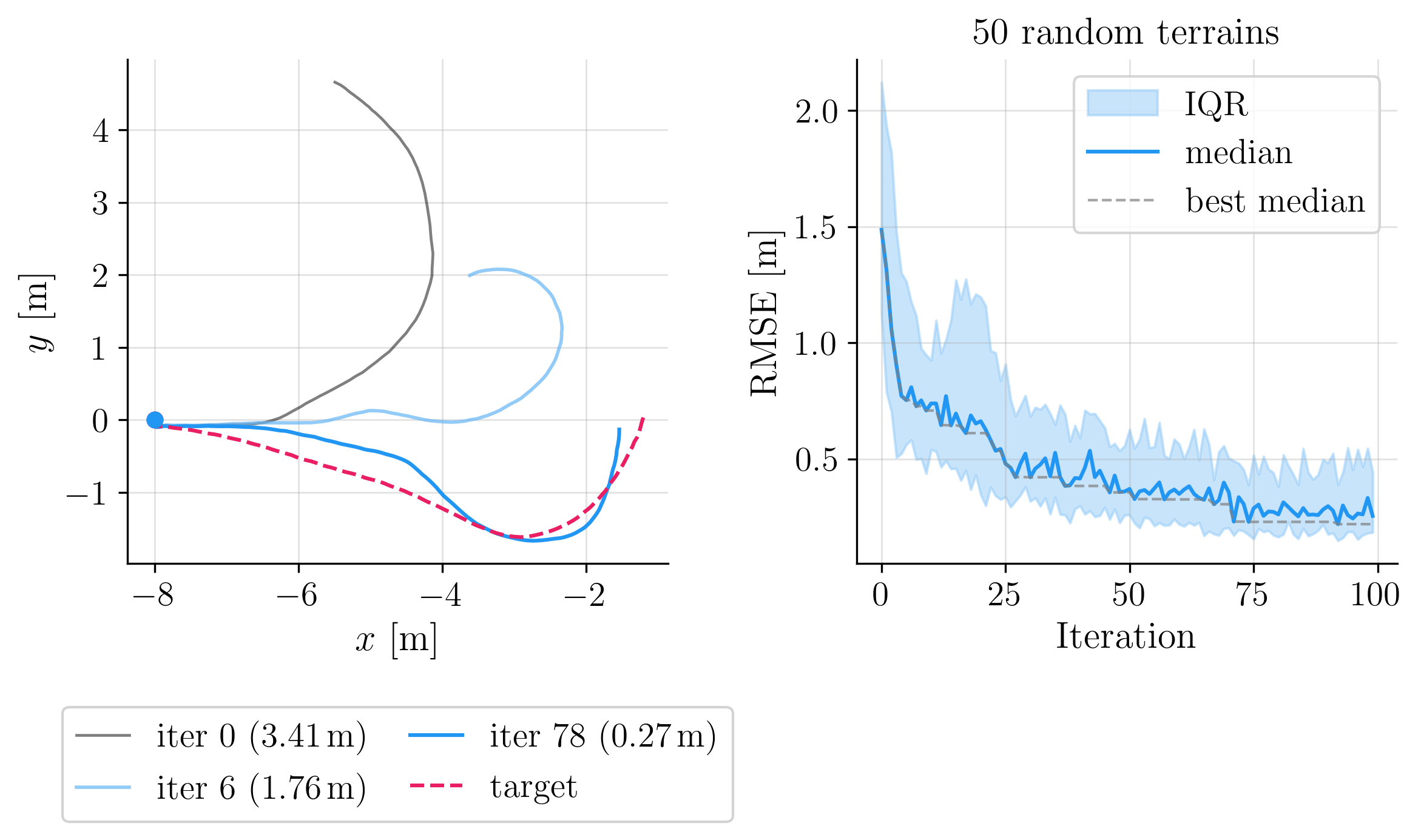}
  \caption{%
    \textbf{Terrain traversal} ($10\,\text{s}$ horizon, $K{=}10$ knots, triangle mesh).
    Left: example seed with initial guess (gray), intermediate iteration (light blue), and best iterate (blue) converging toward the target (dashed red). Right: RMSE median and IQR over $50$ random terrains, $100$ iterations; dashed: running best of the median (the text reports the median over seeds of each seed's best RMSE).~\rev{R1.11}
  }
  \label{fig:terrain}
\end{figure}

\paragraph{Results.}
Fig.~\ref{fig:terrain} shows an example seed and the convergence across all $50$ seeds.~\rev{SP} The per-seed best RMSE has median $\mathbf{0.143\,\text{m}}$ ($0.208 \pm 0.178\,\text{m}$ mean $\pm$ std), with $90\%$ of seeds ($45/50$) below $0.5\,\text{m}$ and $70\%$ ($35/50$) below $0.2\,\text{m}$. The median per-iteration time is $1045\,\text{ms}$ (${\sim}8\,\text{ms/step}$).~\rev{SP}\rev{R1.11}

\section{CONCLUSION}
\label{sec:conclusion}

Ostrich solves a single non-smooth Newton system per timestep that couples hard
contacts, friction, motor control, and articulated dynamics, then differentiates
the converged residual via the implicit function theorem with the forward Schur
complement reused for the adjoint, making per-timestep gradient memory constant
in the inner solver iteration count.~\rev{SP} In Section~\ref{sec:experiments} this delivers MuJoCo's sim-to-real accuracy up to a $\mathbf{50{\times}}$ larger timestep,~\rev{R2.2}\rev{R1.12}
$\mathbf{100\%}$ success on a random-target control-synthesis task where MJX and
Semi-Implicit reach $16\%$ and $8\%$,~\rev{R1.11}\rev{C1} and $\mathbf{8{,}192}$
differentiable parallel worlds on a single $24\,\text{GB}$ GPU where checkpointed MJX reaches $4{,}096$ and Semi-Implicit $512$.~\rev{R2.1} Together these open long horizons, mesh-resolved contact (\S\ref{sec:terrain}), and thousands of parallel worlds to gradient-based optimization.~\rev{SP}

\paragraph{Limitations and future work.}
The implicit Newton solve has a higher per-step cost than explicit integrators,
so Ostrich's per-iteration advantage shrinks when small timesteps are not a
bottleneck. Deformable bodies, not yet supported, could reuse the same Schur-complement backward pass.~\rev{SP} Impacts are resolved inelastically
($e{=}0$); restitution is left to future work.~\rev{R1.10}\rev{R1.gen} All real-robot
evidence comes from a single wheeled platform.~\rev{R2.2} Our adjoint fixes the active set within each step but lets it evolve between steps, so per-step gradients stay valid as the wheels meet new mesh faces; gradient flow through a contact's birth or death, which grasping needs but wheeled locomotion on persistent contact does not, remains open.~\rev{SP}\rev{R1.gen} On long chaotic horizons we truncate the position pull-back (\S\ref{sec:terrain}).~\rev{R1.11}

The architecture targets differentiable MPC on real
hardware~\cite{jahncke2026diffmpc}, population-scale policy learning (RL-style training), and physical-parameter identification via
$\partial\mathbf{r}/\partial\boldsymbol{\theta}$. Code and the robot model will be released.~\rev{SP}

%%%%%%%%%%%%%%%%%%%%%%%%%%%%%%%%%%%%%%%%%%%%%%%%%%%%%%%%%%%%%%%%%%%%%%%%%%%%%%%%

\bibliographystyle{IEEEtran}
\bibliography{references}

\end{document}